\documentclass{article}

\usepackage[final]{neurips_2021}

\usepackage[utf8]{inputenc} 
\usepackage[T1]{fontenc}    
\usepackage{hyperref}       
\usepackage{url}            
\usepackage{booktabs}       
\usepackage{amsfonts}       
\usepackage{nicefrac}       
\usepackage{microtype}      
\usepackage{xcolor}         

\usepackage{graphicx}
\usepackage{wrapfig}
\usepackage{subcaption}
\usepackage{multirow}

\title{Deeply Shared Filter Bases for 
Parameter-Efficient 
Convolutional Neural Networks}

\author{%
  Woochul~Kang \\
  Department of Embedded Systems Engineering\\
  Incheon National University\\
   Yeonsu-gu, South Korea 22012 \\
  \texttt{wchkang@inu.ac.kr} \\
   \And
    Daeyeon~Kim\thanks{W. Kang and D. Kim are contributed equally. This work was conducted at Incheon National University. }\\
  Dain Technology, Inc. \\
  \texttt{dustin@soundablehealth.com} \\
}

\begin{document}

\maketitle

\begin{abstract}
Modern convolutional neural networks (CNNs) have massive identical convolution blocks, 
and, hence, recursive sharing of parameters across these blocks
has been proposed to reduce the amount of parameters.
However, naive sharing of parameters poses many challenges such as 
limited representational power
and the vanishing/exploding gradients problem of recursively shared parameters.
In this paper, 
we present a recursive convolution block design and training method,
in which a recursively shareable part, or a filter basis, 
is separated and learned while effectively avoiding the vanishing/exploding gradients problem during training.
We show that the unwieldy vanishing/exploding gradients problem
can be controlled by enforcing the elements of the filter basis orthonormal,
and empirically demonstrate that
the proposed \emph{orthogonality regularization} 
improves the flow of gradients during training.
Experimental results on image classification and object detection 
show that our approach, unlike previous parameter-sharing approaches, 
does not trade performance to save parameters and 
consistently outperforms overparameterized counterpart networks.
This superior performance demonstrates that the proposed recursive convolution block design and the orthogonality regularization
not only prevent performance degradation, 
but also consistently improve
the representation capability  
while a significant amount of parameters are recursively shared.
\end{abstract}

\section{Introduction}

Modern convolutional neural networks (CNNs) such as ResNets 
have massive identical convolution blocks
and 
recent analytic studies \citep{jastrzebski2018residual} show  
that these blocks perform mostly iterative refinement of features
rather than learning new features.
Inspired by these results, 
recursive sharing of weights 
has been studied as a promising direction to parameter-efficient CNNs
\citep{jastrzebski2018residual,guo2019dynamic,savarese2019learning}.
However, naive sharing of parameters across many convolution layers
incurs several problems. 
First of all, recursive sharing of parameters can result in 
the \textit{vanishing} and the \textit{exploding gradients} problem,
which is one of the main reasons that recurrent neural networks (RNNs)
are so hard to train properly
\citep{pascanu2013difficulty, jastrzebski2018residual}.
Another problem 
is that 
overall representation power can be limited 
by iterative sharing of parameters.
Due to these challenges, 
most compression approaches based on parameter-sharing 
suffer from performance degradation.

In this work, we conjecture that convolution layers or blocks 
can be separated into inherently shareable parts and non-shareable parts,
and can be trained effectively by avoiding the vanishing and the exploding gradients problem.
To achieve this, for a full convolution operator, we first replace it 
with a factorized version that splits the convolution operator into two separate operators;
one operator with inherently shareable filters, called a \emph{filter basis}, 
and the other operator with non-shared filters, called \emph{coefficients}. 
When successive convolution blocks 
share a common filter basis,
they are positioned in the same vector subspace.
However, their representation capability is retained
through non-shared coefficients
that learn diverse features by linearly combining the shared filter basis.  

By separating shareable parts from non-shareable parts, 
we can impose desirable properties on the shared parameters.
To avoid performance degradation from recursive sharing of parameters,
we propose the \emph{orthogonality regularization},
in which the vanishing/exploding gradients problem is controlled 
by enforcing the elements of a shared filter basis orthonormal during 
training.
We both theoretically and empirically 
show that the proposed orthogonality regularization 
improves the flow of the gradients during training
and reduces the redundancy in parameters effectively.

For efficient CNNs such as MobileNets \citep{howard2017mobilenets},
we do not need to factorize convolution operators
to uncover a shared filter basis
since these networks already have 
factorized convolution block structures for computational efficiency.
For such networks, 
our approach can be applied 
simply by identifying one or two convolution operators of repeating 
convolution blocks as a filter basis that shares weights across the repeating blocks.
Other convolution operators in each convolution block become block-specific 
non-shared coefficients.



Since our focus is not on pushing the state-of-the-art performance, 
we demonstrate the effectiveness of our work using widely-used models 
as base models
on image classification and object detection.
Without bells and whistles, 
simply applying
the proposed convolution block design and the orthogonality regularization
saves a significant amount of parameters 
while consistently outperforming over-parameterized counterpart networks. 
%
For example, our method can save up to 46.0\% of parameters of ResNets
while consistently achieving lower test errors. 
Even in compact models, such as MobileNetV2, our approach can achieve 
further 8-21\%  parameter savings while outperforming the original models.
This superior performance demonstrates that 
the proposed recursive convolution blocks and orthogonality regularization
enables effective learning 
of better feature representations while a significant amount of parameters
are shared recursively.



\section{Related Work}
\textbf{Recursive networks and parameter sharing:} 
Recurrent neural networks (RNNs) \citep{graves2013speech} have been well-studied for temporal
and sequential data. 
As a generalization of RNNs, recursive variants of CNNs 
are used extensively for visual tasks
\citep{socher2011parsing,liang2015recurrent,xingjian2015convolutional,kim2016deeply,zamir2017feedback}.
For instance, \citet{eigen2014understanding} explore recursive convolutional architectures that share filters across multiple convolution layers. 
They show that recurrence with deeper layers tends to increase performance. 
However, their recursive architecture shows worse performance than independent 
convolution layers due to overfitting. 
In contrast, we share only fundamentally shareable parts, or filter bases,
that are separated from conventional convolution operators. 
By separating shareable parts from non-shareable parts, 
we can impose desirable properties on the shared parameters that 
prevents the vanishing/exploding gradients problems
without damaging representational capability of the networks. 

More recently, \citet{jastrzebski2018residual} show that 
iterative refinement of features in ResNets suggests that deep networks can potentially 
leverage intensive parameter sharing. 
\citet{guo2019dynamic} introduce a gate unit to determine whether to jump out of the 
recursive loop of convolution blocks to save computational resources.
These works show that training recursive networks with naively shared blocks leads to bad performance
due to the problem of gradient explosion and vanish like RNN \citep{pascanu2013difficulty,pmlr-v70-vorontsov17a}.
In order to deal with the problem of gradient explosion and vanish,
they suggest unshared batch normalization strategy.
In our work, we propose the orthogonality regularization of shared filter bases
to further address this problem. 

A few recent works generate convolution filters 
by combining shared building blocks \citep{bhalgat2020structured, savarese2019learning, yang2019legonet, qiu2018dcfnet}. 
Although these approaches are different in details, they are similar to our work since they all share 
a set of filters 
and combine them to build layer-specific filters. 
However, they save parameters at the cost of accuracy loss. 
For example, in ResNet50 on ImageNet, \citet{yang2019legonet}'s work incurs 0.9\% accuracy drop, 
while our method achieves performance improvement. 
We believe that this performance gap comes from the gradients issue of shared parameters 
and these previous works could benefit more from the improved flow of gradients of our approach.



\textbf{Model compression and efficient convolution block design:} 
Reducing storage and inference time of CNNs has 
been an important research topic for both
resource constrained mobile/embedded systems and energy-hungry data centers.
A number of research techniques have been developed 
such as 
filter pruning \citep{lecun1990optimal,polyak2015channel,li2017pruning,he2017channel}, 
low-rank factorization \citep{denton2014exploiting,jaderberg2014speeding},
and quantization \citep{han2016deep},
to name a few. 
Several model compression techniques 
factorize trained filters to reduce the computation complexity 
\citep{zhang2015accelerating, li2019learning}. 
Our proposed block structure also 
benefit from the reduced computational complexity too by factorizing 
convolution filters.
However, most previous compression techniques have been suggested as post-processing steps that are applied after initial training.
Therefore, their accuracy is usually bounded by the approximated
base models.  
In contrast,  our primary goal in 
factorizing convolution operators is to 
find a common filter basis shared by iterative convolution layers,
so the weights of the shared filter basis are 
learned from scratch, not from decomposing pretrained filters.
Hence, 
unlike previous compression techniques, 
the performance of our approach is not limited
by the original models,
and the experimental results show that 
our parameter-sharing approach consistently 
outperforms the overparameterized counterpart networks
on various datasets and tasks. 






\section{Proposed Method}

\begin{figure*}[tbp]
\centering
  \includegraphics[width=1.0\linewidth]{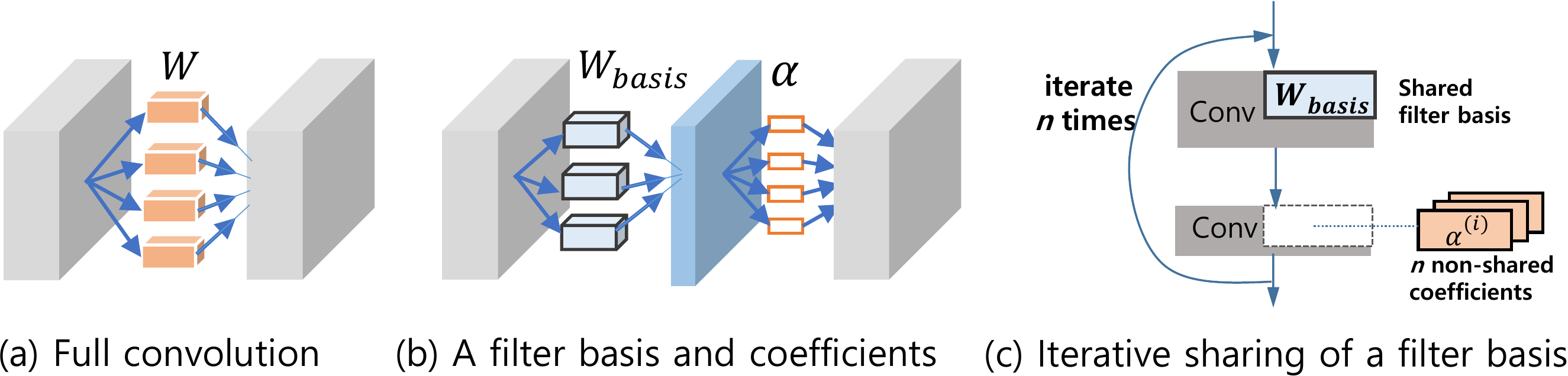}
  \caption{
  A full convolution operator in (a) with a filter $W$ can be replaced by 
  two separate convolution operators, as in (b): one operator with a \emph{filter basis} ($W_{basis}$) and the other operator with \emph{coefficients} ($\alpha$s). 
  A filter basis is inherently shareable for convolution operators
  in the same vector subspace, and,
  hence, a recursive architecture is suggested, as in (c).
  The filter basis shared by recursive convolution layers
  can be learned through typical gradient-based training
  while avoiding potential vanishing/exploding gradients problem 
  by enforcing its elements orthonormal.
  }
  \label{fig:conv}
\end{figure*}

In this section, we describe the design of recursive convolution blocks
and discuss how to train them 
to deal with the vanishing/exploding gradients problem.  

\subsection{Recursive Sharing of a Filter Basis}
\label{sec:method_filterbasis}
Naive sharing of convolution filters across repetitive layers 
degrades the overall performance due 
to limited representation power of individual layers.
We argue that convolution filters can be 
separated into fundamentally shareable components
and non-shareable components, 
and the performance degradation can 
be prevented by 
sharing only fundamentally shareable parts.


More formally, we consider a convolution layer, shown in Figure \ref{fig:conv} (a), that has $S$ input channels, $T$ output channels, and a set of filters $W = \{W_t\ \in R ^{k \times k \times S}, t \in [1..T]\}$.
Each filter $W_t$ can be decomposed into a filter basis $W_{basis}$
and coefficients $\mathbf{\alpha}$:
\begin{equation}
    \centering
    W_t =  \sum_{r=1}^R \alpha^r_{t} W^r_{basis},   
    \label{eq:decomp}
\end{equation}
where $W_{basis}= \{W^r_{basis} \in \mathbb{R}^{k \times k \times S}, r \in [1..R] \}$ 
is a filter basis, and 
$\mathbf{\alpha}=\{\alpha_t^r \in \mathbb{R}, r \in [1..R], t \in [1..T]\}$ 
is scalar coefficients. 
In Equation \ref{eq:decomp}, $R$ is the rank of the basis.
In a typical convolution layer, output feature maps $V_t \in \mathbb{R}^{w \times h \times T}, t \in [1..T]$
are obtained by the convolution 
between input feature maps $U \in \mathbb{R}^{w \times h \times S}$ 
and filters $W_t, t \in [1..T]$. 
With Equation \ref{eq:decomp}, this convolution can be rewritten as follows:
\begin{equation}
    V_t  = U \ast W_t  = U *  \sum_{r=1}^R \alpha^r_t W^r_{basis} \label{eq:decompconv-1} 
       = \sum_{r=1}^R \alpha^r_t ( U * W^r_{basis}), \textrm{ where} \ t \in [1..T]. 
\end{equation}
In Equation \ref{eq:decompconv-1}, the order of the convolution operation 
and the linear combination of filter basis is reordered according to the linearity of convolution operators. 
Therefore, a conventional convolution layer 
can be replaced with two successive convolution layers as shown in Figure \ref{fig:conv}-(b).

This factorized convolution filters suggest 
a recursive architecture, shown in Figure \ref{fig:conv}-(c),
in which a common filter basis is used for 
iterative convolution layers.
Since all filters in the recursive loop 
have a common filter basis, they 
are positioned in the same vector subspace.
However, their representation capability is 
still retained since their coordinates in the subspace 
are diversified by their respective non-shared coefficients.



\subsection{Orthonormality of Shared Filter Bases}
\label{sec:method_proof}

For recursive sharing of a filter basis, as in Figure \ref{fig:conv}-(c),
we need to find an optimal filter basis that 
can be used by iterative convolution layers without loss of performance.
Although this optimal filter basis can be searched by typical
gradient-based optimization such as SGD, 
one major problem is that the exploding/vanishing gradients problem of
recursively shared filter bases
can prevent proper search of optimization space \citep{pascanu2013difficulty}. 

More formally, we consider a series of $N$ factorized convolution blocks,
in which a filter basis $W_{basis}$ is shared $N$ times,
as in Figure \ref{fig:conv}-(c).
Let $\mathbf{x}^i$ be the input of the $i$-th convolution block,
and $a^{i+1}$ be the output of the 
convolution of $\mathbf{x}^{i}$ with the filter basis $W_{basis}$
\begin{equation}
    \centering
    a^{i}(\mathbf{x}^{i-1}) = W_{basis}^\top  \mathbf{x}^{i-1}.
    \label{eq:shared-conv}
\end{equation}

In Equation \ref{eq:shared-conv},
$W_{basis} \in \mathbb{R}^{k^2S \times R}$ is a reshaped filter basis 
that has basis elements at its columns.
We assume that input $\textbf{x}$ is properly adapted 
(e.g., with \textit{im2col} operations)
to express convolutions using a matrix-matrix multiplication.
Since $W_{basis}$ is shared by $N$ recursive convolution blocks,
the gradient of $W_{basis}$ for some loss function $L$ is:
\begin{equation}
    \centering
    \frac{\partial L}{\partial W_{basis}} = 
    \sum_{i=1}^{N}
    \frac{\partial L}{\partial a^N} \prod_{j=i}^{N-1}\left(
            \frac{\partial a^{j+1}}{\partial a^{j}} \right)
    \frac{\partial a^i}{\partial W_{basis}},
    \label{eq:grad-loss}
\end{equation}
, where
\begin{equation}
    \centering
    \frac{\partial a^{j+1}}{\partial a^{j}} = 
    \frac{\partial a^{j+1}}{\partial \mathbf{x}^{j}} \frac{\partial \mathbf{x}^{j}}{\partial a^{j}} =
    W_{basis} \frac{\partial \mathbf{x}^{j}}{\partial a^{j}}
    \label{eq:grad-state}
\end{equation}

If we plug $W_{basis} \frac{\partial \mathbf{x}^{j}}{\partial a^{j}}$ in Equation \ref{eq:grad-state} 
into Equation \ref{eq:grad-loss}, 
we can see that $\prod \frac{\partial a^{j+1}}{\partial a^{j}}$ 
is the term that makes gradients unstable
since $W_{basis}$ is multiplied many times. 
This exploding/vanishing gradients can be controlled to a large extent by keeping $W_{basis}$ 
close to orthogonal \citep{pmlr-v70-vorontsov17a}.
For instance, if $W_{basis}$ admits eigendecomposition, $[W_{basis}]^N$ can 
be rewritten as follows:
\begin{equation}
    \centering
    [W_{basis}]^N = [Q \Lambda Q^{-1}]^N = Q \Lambda^N Q^{-1},
    \label{eq:w-times}
\end{equation}
where $\Lambda$ is a diagonal matrix with the eigenvalues placed on the diagonal
and $Q$ is a matrix composed of the corresponding eigenvectors.
If $W_{basis}$ is orthogonal, $[W_{basis}]^N$ neither explodes nor vanishes,
since all the eigenvalues of an orthogonal matrix have absolute value of 1.
This result shows that a shared filter basis with orthonormal elements
ensures that forward and backward signals neither explode nor vanish.

Based on this result, 
we propose the \emph{orthogonality regularization}
\footnote{Technically, the columns (or rows) of an orthogonal matrix form
an \emph{orthonormal basis}. 
However, we use the term \emph{orthogonality regularization} 
since orthonormal bases have no proper term for their matrix-form.}
to enforce orthonormality to shared filter bases during training.
For instance, when convolution operators 
in each residual block group of a ResNet shares a filter basis, 
the objective function $L_R$ can be defined
to have the orthogonality regularization term in addition
to the original loss $L$:
\begin{equation}
\centering
L_{R} = L + \lambda \sum_g^G{\| W_{basis}^{(g)}{}^\top \cdot W_{basis}^{(g)} - I \|^2},
\label{eq:train-loss}
\end{equation}
where $W_{basis}^{(g)}$ is a shared filter basis  for $g$-th residual block group and 
$\lambda$ is a hyperparameter. 

In Equations \ref{eq:grad-loss} and \ref{eq:grad-state}, 
we also need to ensure that the norm of $\frac{\partial \mathbf{x}^{j}}{\partial a^{j}}$ is bounded for stability during forward and backward passes \citep{pascanu2013difficulty}.
It is shown that batch normalization after non-linear activation at each convolution layer ensures healthy norms \citep{ioffe2015batch,guo2019dynamic,jastrzebski2018residual}.
In Section \ref{sec:analysis-ortho},
we empirically show that the proposed orthogonality regularization 
and batch normalization similarly improve 
the flow of gradients during training.

\subsection{Enhancing Representation Capability}
\begin{figure}[tbp]
\centering
  \includegraphics[width=1.0\linewidth]{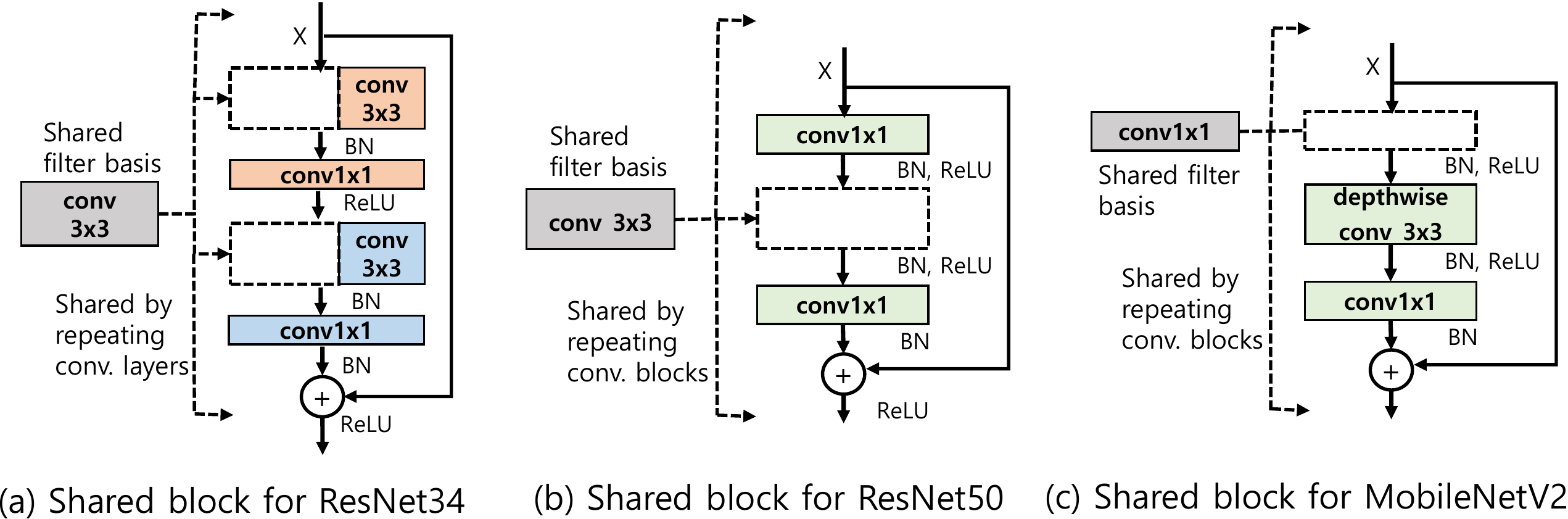}
  \caption{The structure of residual blocks with a shared filter basis.
  For ResNet34, two full convolution operators are factorized to uncover a shared filter basis.
  For ResNets with bottleneck blocks (ResNet50/101) and MobileNetV2, 
  one or two computationally expensive convolution operators of each convolution block
  are selected as shared filter bases and the other convolution operators
  are designated as non-shared coefficients.
  In (a), the filter basis has non-shared elements to further increase representation capability. 
 }
  \label{fig:blockarch}
\end{figure}

When convolution operators  
share a common filter basis, 
they are all in the same vector subspace.
Therefore, if the rank of the filter basis is too low, 
it might limit the representation capability of the 
convolution operators sharing the filter basis.
Conversely, if the rank of the shared filter basis is too high (e.g., $R \ge T$),
the computational gain of factorized structure is mitigated. 
One way to increase the representational power of each convolution operator, 
while still maintaining its computational complexity low,
is placing the convolution operators in different 
subspace by adding a small number of non-shared elements to the filter basis. 
For instance, we build a filter basis $W_{basis}$ by combining shared elements 
and non-shared elements:
\begin{equation}
\centering
W_{basis}= 
W_{bs\_shared} \cup W_{bs\_unique}, 
\label{eq:basis}
\end{equation}
where $W_{bs\_shared}= \{W^{r}_{bs\_shared} \in \mathbb{R}^{k \times k \times S}, r \in [1..n] \}$
are shared filter basis elements, and 
$W_{bs\_unique}= \{W^{r}_{bs\_unique} \in \mathbb{R}^{k \times k \times S}, r \in [n{+}1..R] \}$ 
are non-shared filter basis elements.
For example, Figure \ref{fig:blockarch}-(a) shows that 
the filter basis of ResNet34 
is compose of both shared and non-shared elements. 
One disadvantage of this hybrid scheme is 
that non-shared filter basis elements require more parameters. 
The ratio of non-shared basis elements can be varied to control the tradeoffs. 
But, our results in Section \ref{sec:analysis-rank} show that providing only a few 
non-shared elements to a filter basis is enough to achieve high performance.

\section{Experiments}
\label{sec:experi}
In this section, we perform a series of experiments on image classification 
and object detection using several modern networks as base models.
We also analyze the effect of the orthogonality regularization.

For ResNets with conventional convolution filters (e.g, ResNet34), 
a filter is replaced by the proposed factorized convolution block
as shown in Figure \ref{fig:blockarch}-(a).
A filter basis is shared within a residual block group having 
same kernel dimensions. 
Throughout the experiments, 
we denote by ResNet$L$-S$s$U$u$ a ResNet with $L$ layers
that has a filter basis with a $s$-to-$u$ ratio 
of shared elements and non-shared elements.

ResNets with bottleneck blocks (e.g., ResNet50)
and MobileNetV2 already have decomposed block structures.
Therefore, 
we designate one or two convolution operators
with the largest parameters 
in each block as filter bases sharing weights across 
iterative blocks,
as shown in Figure \ref{fig:blockarch}-(b) and -(c).
During the training, the proposed orthogonality regularization is applied 
to enforce the elements of the shared filter bases orthonormal.
In these models, we leave at least one convolution operator as non-shared coefficients to improve expression ability. 
In Section \ref{section_ilsvrc}, we explore the effect of sharing both a filter basis and coefficients in residual groups.

During experiments, 
all programs for training and evaluation run on PCs equipped 
either with four RTX 2080Ti GPUs or two RTX 3090 GPUs and 
an Intel i9-10900X CPU @3.7GHz. 

\subsection{Image Classification on ImageNet} 
\label{section_ilsvrc}

We evaluate our method on the ILSVRC2012 dataset \citep{russakovsky2015imagenet} that has 1000 classes. 
The dataset consists of 1.28M training and 50K validation images. 
We use ResNets and MobileNetV2 as base models.
We train the ResNet-derived models for 150 epochs with SGD optimizer with a mini-batch size of 256, 
a weight decay of 1e-4, and a momentum of 0.9. 
The learning rate starts with 0.1 and decays by 0.1 at 60-th, 100-th, and 140-th epochs.
MobileNetV2 and our MobileNetV2-Shared models are trained 
for 300 epochs with a weight decay of 1e-5.
Its learning rate starts with 0.1 and decays by 0.1 at 150-th, 225-th, and 285-th epochs.

\begin{table*}[tbhp!]
\centering
\begin{footnotesize}
  \caption{Error (\%) on ImageNet. 
  In ResNet50/101-Shared$^{\ddagger}$ and MobileNetV2-Shared$^{\ddagger}$, 
  first two convolution operators of each residual block 
  are designated as recursively shared filter bases.
  In ResNet50-Shared-All, both a filter basis and coefficients
  are shared recursively across blocks of each residual group.
  In ResNet50-Shared-NoOrthoReg, orthogonality regulaization is not applied to ResNet50-Shared
  during training.
  Latency is measured on the Nvidia Jetson TX2 embedded board (GPU, batch size = 1). }
  \label{table:ilsvrceval}
  \begin{tabular}[htbp]{l|llllll}
    \toprule
    Baseline & Model     & Params     & FLOPs & top-1 & top-5 & Latency  \\
    \midrule
    \multirow{3}{*}{ResNet34} & ResNet34 (baseline) &  21.80M  & 3.67G & 26.70 & 8.58  & 33.6ms\\ 
    & Filter Pruning \citep{li2017pruning} & 19.30M & 2.76G & 27.83 & - &-\\
    & ResNet34-S48U1 (ours) & 11.79M & 3.26G & \textbf{26.67} & \textbf{8.54} & 38.6ms\\ 
       \midrule
       \multirow{7}{*}{ResNet50} & ResNet50 (baseline) &  25.56M  & 4.11G & 23.85 & 7.13  & 43.8ms\\ 
      & Versatile-ResNet50 \citep{wang2018learning} &  19.0M  & 3.2G & 24.5 & 7.6  & -\\ 
      & FSNet \citep{yang2020fsnet} &  13.9M  & - & 26.89 & 8.63  & -\\ 
    & ResNet50-Shared (ours) & 20.51M & 4.11G & \textbf{23.64} & \textbf{6.98}  & 43.3ms\\  
    & ResNet50-Shared$^\ddagger$ (ours) & 18.26M & 4.11G & 23.95 & 7.14  & 43.3ms\\  
    & ResNet50-Shared-All & 16.02M & 4.11G & 24.35 & 7.41  & -\\  
    & ResNet50-Shared-NoOrthoReg & 20.51M & 4.11G & 24.19 & 7.34  & -\\  
    
     \midrule
    \multirow{2}{*}{ResNet101} & ResNet101 (baseline) &  44.55M  & 7.83G & 22.63 & 6.44 &  73.2ms\\
    & ResNet101-Shared (ours)  & 29.47M & 7.83G & \textbf{22.31} &  6.47 & 72.9ms \\
    \midrule
    \multirow{4}{*}{MobileNetV2} & MobileNetV2 (baseline) &  3.50M  & 0.33G & 28.0 & 9.71 &  18.4ms\\
    & DR-MobileNetV2 \citep{guo2019dynamic}  & 2.96M & 0.27G & 28.2 & 9.72 & - \\ 
        & MobileNetV2-Shared (ours)  & 3.24M & 0.33G & \textbf{27.61} &  \textbf{9.34} & 17.9ms \\
    & MobileNetV2-Shared$^{\ddagger}$ (ours)  & 2.98M & 0.33G & 28.21 &  9.85 & 17.8ms \\
    \bottomrule
  \end{tabular}
\end{footnotesize}
\end{table*}

In Table \ref{table:ilsvrceval}, we compare our results with competing techniques.
Most parameter-sharing and compression techniques save 
parameters at the cost of performance.
Unlikely, the results in Table \ref{table:ilsvrceval} show that 
our approach consistently outperforms the counterpart networks
while saving a significant amount of parameters.
For example, ResNet34-S48U1 
outperforms the counterpart ResNet34 while only using 54.0\% parameters.
Since our models derived from ResNet50/101 and MobileNetV2
already have factorized convolution blocks,
their overall parameter-saving is not as high as ResNet34-S48U1's.
However, they still save about 19.8\%, 33.9\% and 7.5\% parameters, respectively,
while outperforming the baselines.
The benefit of our method is more pronounced in deeper networks. 
In ResNet101, for example, 
a filter basis is shared by up to 23 recursive convolution layers,
resulting in 35.7\% reduction of parameters.
In Table \ref{table:ilsvrceval},
ResNet50-Shared-NoOrhtoReg shows the effect of not applying 
the orthogonality regularization on ResNet50-Shared. 
Our result shows that its top1 accuracy drops to 75.81\%, 
that is 0.34\% lower than the counterpart ResNet50. 
In contrast, when the orthogonality regularization is applied, the same model achieves 76.35\%, 
which is 0.21\% higher than the counterpart ResNet50. 
This 0.55\% improvement is obtained simply by applying orthogonality regularization. 
In Table \ref{table:ilsvrceval},
ResNet50-Shared-All shows the effect of sharing not only a filter basis but also coefficients
recursively within residual groups. 
The result shows that we can save further 18\% parameters (reduced from 20.51M to 16.02M) of ResNet50-Shared, 
but its top-1 (top-5) accuracy drops to 75.65\% (92.59\%), which is 0.5\% lower than the counterpart ResNet50. 
This result shows that having unshared coefficients is important for high performance.
\citet{yang2019legonet} also demonstrates that sharing coefficients drops the accuracy by additional 2\%. 

Although ResNet34-S48U1 requires lower FLOPs than the counterpart ResNet34, 
it takes 14\% longer latency on Jetson TX2. 
This overhead mostly comes from separated convolution operations
for shared and non-shared filter basis elements.
Current GPU-based deep learning libraries are not optimized to process such 
separated operations efficiently. 
Unlikely, 
ResNet50/101-Shared and MobileNetV2-Shared 
do not have such separated convolution operations,
and, hence, their latency is slightly lower on the device
that is constrained by limited cache and memory. 

\subsection{Image Classification on CIFAR} \label{section_CIFAR}

We evaluate the effectiveness of our method on various modern CNNs 
with the CIFAR dataset that has 50,000 and 10,000 
$32\times32$ images for training and testing, respectively.  
For training networks, 
we follow a similar training scheme in \citet{he2016deep}.
Standardized data-augmentation and normalization are applied to input data. 
Networks are trained for 300 epochs with SGD  optimizer 
with a weight decay of 5e-4 and a momentum of 0.9. 
The learning rate is initialized to 0.1 and 
is decayed by 0.1 at 50\% and 75\% of the epochs. 

\begin{table*}[tbhp!]
\centering
\begin{footnotesize}
  \caption{Error (\%) on CIFAR-100. `$^\star$' denotes that the orthogonality regularization is not applied.}
  \label{table:cifar100eval}
  \begin{tabular}[htbp]{l|lllll}
    \toprule
    Baseline & Model     & Params     & FLOPs & Error \\
    \midrule
    \multirow{3}{*}{ResNet34}& ResNet34 (baseline)  & 21.33M & 1.17G & 22.49 \\ 
            & ResNet34-S32U1$^\star$ (ours) & 7.73M & 0.78G & 22.92 \\
            & ResNet34-S32U1 (ours) & 7.73M & 0.78G & \textbf{21.84} \\
    \midrule
    \multirow{2}{*}{DenseNet121} & DenseNet121 (baseline) & 7.05M & 0.91G & 21.95  \\ 
            & DenseNet121-S16U1 (ours) & 5.08M & 0.72G & 22.15 \\
    \midrule
    \multirow{2}{*}{ResNeXt50} & ResNeXt50 (baseline) & 23.17M & 1.36G & 20.71 \\ 
            & ResNeXt50-S64U4 (ours) & 19.3M & 1.19G & \textbf{20.09}\\
    \bottomrule
  \end{tabular}
\end{footnotesize}
\end{table*}

\begin{table*}[tbhp!]
\centering
\begin{footnotesize}
  \caption{Error (\%) on CIFAR-10. `$^\ddagger$' denotes having 2 shared bases 
  in each convolution block group. `$\star$' denotes that the orthogonality regularization is not applied.}
  \label{table:cifar10eval}
  \begin{tabular}[htbp]{l|lllll}
    \toprule
    Baseline & Model     & Params     & FLOPs & Error \\
    \midrule
    \multirow{2}{*}{ResNet32} & ResNet32 (baseline)  & 0.46M & 0.07G & 7.51  \\ 
            & ResNet32-S16U1$^\ddagger$ (ours) & 0.24M & 0.08G & \textbf{6.95}\\
    \midrule
    \multirow{5}{*}{ResNet56}& ResNet56 (baseline) & 0.85M & 0.16G & 6.97  \\ 
            &Filter Pruning \citep{li2017pruning} & 0.77M  & 0.09G &6.94\\
            &KSE \citep{li2019exploiting} & 0.43M & 0.06G & 6.77\\
            &DR-Res 40 \citep{guo2019dynamic} & 0.50M  & 0.11G &6.51\\
            & ResNet56-S16U1$^\star$ (ours) & 0.27M & 0.15G & 7.70\\  
            & ResNet56-S16U1 (ours) & 0.27M & 0.15G & 7.46\\  
            &ResNet56-S16U1$^\ddagger$ (ours) & 0.31M & 0.15G & \textbf{6.33}\\ 
    \bottomrule
  \end{tabular}
\end{footnotesize}
\end{table*}

Table \ref{table:cifar100eval} shows the results on 
CIFAR-100.
Networks trained with the proposed method consistently outperform their counterparts.
For instance, ResNet34-S32U1 requires only 36.2\% parameters and 66.6\% FLOPs of the counterpart ResNet34
while achieving lower test error (21.84\%) than much deeper ResNet50 (22.36\%).
To show the generality of our work, we apply the proposed method
to DenseNet \citep{huang2017densely}, and ResNeXt \citep{xie2017aggregated}.
Although the overall gain is not as pronounced as ResNets',
we still observe reduction of resource usages in these networks. 
For instance, ResNeXt50-S64U4 outperforms the counterpart ResNeXt50 
while saving parameters and FLOPs by 16.7\% and 12.1\%, respectively.
In ResNeXt, the gain is limited since they mainly exploit group convolutions;
each group convolution is decomposed for filter basis sharing in our network.

The result on CIFAR-10 with ResNets is presented in Table \ref{table:cifar10eval}.
Unlike networks on CIFAR-100, 
networks on CIFAR-10 has much fewer channels (e.g. 16 channels in the first residual block group)
and, hence, projecting filters to 
such low dimensional subspace might limit the performance of the networks. 
For instance, in ResNet32-S8U1, 
filters are supposed to be projected 
onto 9 dimensional subspace consisting of 8 shared and 1 non-shared filter basis elements. 
Further, for deeper networks such as ResNet56, 
a filter basis is supposed to be shared by many residual blocks in the group, 
and it can damage the performance. 
For example, every filter basis in ResNet56-S16U1 is shared by 
8 residual blocks, or 16 convolution operators.
Due to this excessive sharing, though ResNet56-S16U1 saves 41.3\% parameters,
its testing error (7.46\%) is 
higher than the counterpart ResNet56's (6.97\%).

To remedy this problem, we introduce a variant, in which 
each residual block group of the networks uses 2 shared bases; 
one basis is shared by the first convolution operators of recursive residual blocks,
and the other basis is shared by the second convolution operators. 
In Table \ref{table:cifar10eval}, networks with a `$^\ddagger$' mark denote this variant.
Though this variant slightly increases the parameters of the networks,
it can prevent excessive sharing of parameters. 
For example, although ResNet56-S16U1$^\ddagger$
needs 0.04M more parameters for additional shared bases, 
it still saves 63\% parameters of the counterpart ResNet56
and achieves lower testing error of 6.33\%.

In Table \ref{table:cifar10eval}, we compare our results with similar state-of-the-art techniques.
Our method achieves better performance and parameter-saving than 
other approaches such as 
filter pruning \citep{li2017pruning}, 
kernel clustering \citep{li2019exploiting}, and recursive parameter sharing \citep{guo2019dynamic}.



\subsection{Object Detection on MS COCO}

In order to explore the generalization ability of our approach,
we next use COCO 2017 dataset 
on object detection task using 
Faster R-CNN \citep{ren2017faster}, 
Mask R-CNN \citep{he207mask}, and RetinaNet \citep{lin2017focal} as detectors.
We compare ResNet50/101 and our ResNet50/101-Shared as backbone networks of the detectors.
These backbone networks are pre-trained on ImageNet, then are transferred 
to MS COCO by fine-tuning. 
We use MMDetection \citep{mmdetection} toolbox and employ default settings 
for training and evaluation. 
All networks are trained on \texttt{train2017} for 12 epochs
using SGD with weight decay of 1e-4, momentum of 0.9 
and mini-batch size of 8 (2 examples per GPU). 
The learning rate is initialized to 0.01 and decays by 0.1 at 8-th and 11-th epochs. 

\begin{table*}[tbhp!]
\centering
\begin{footnotesize}
  \caption{Object detection results on COCO 2017 validation set.
  ResNet50-Shared$^{\ddagger}$ uses first two convolution operators
  of each residual block as recursively shared filter bases. }
  \label{table:coco17detection}
  \begin{tabular}[tbhp]{l | l | l | l | l | l | l | l | l | l }
  \hline 
  Backbone & Detector & \#Params & GFLOPs & $AP$ & $AP_{50}$ & $AP_{75}$ & $AP_{S}$ & $AP_{M}$ & $AP_{L}$ \\
  \hline
  ResNet50 (baseline) & \multirow{5}{4em}{Faster R-CNN}  & 41.53M & 207.07 & 36.4 & 58.2 & 39.2 & 21.8 & 40.0 & 46.2 \\
  -Shared (ours) &  & \textbf{36.70}M & 206.87 & \textbf{37.2} & 58.1 & 40.2 & 21.6 & 40.8 & 47.9  \\
  -Shared$^{\ddagger}$ (ours) &  & \textbf{34.46}M & 206.87 & \textbf{36.6} & 57.4 & 39.8 & 21.4 & 40.1 & 47.3 \\
  \cline{1-1}\cline{3-10}
ResNet101 (baseline) &   & 60.52M & 283.14 & 38.7 & 60.6 & 41.9 & 22.7 & 43.2 & 50.4 \\
-Shared (ours) &  & \textbf{45.67}M & 282.84 & \textbf{39.0} & 59.7 & 42.7 & 22.4 & 42.7 & 50.8  \\
  \hline
  ResNet50 (baseline)& \multirow{5}{4em}{Mask R-CNN} & 44.18M & 275.58 & 37.2 & 58.9 & 40.3 & 22.2 & 40.7 & 48.0 \\
  -Shared (ours)&  & \textbf{39.35}M & 259.94 & \textbf{37.9} & 58.4 & 41.4 & 22.4 & 41.2& 49.2 \\
  -Shared$^{\ddagger}$ (ours)&  & \textbf{37.10}M & 259.94 & \textbf{37.3} & 57.6 & 40.7 & 21.5 & 40.3& 48.5 \\
  \cline{1-1}\cline{3-10}
  ResNet101 (baseline)&  & 67.17M & 351.65 & 39.4 & 60.9 & 43.3 & 23.0 & 43.7 & 51.4 \\
-Shared (ours)&  & \textbf{48.31}M & 335.91& \textbf{39.8} & 60.3 & 43.6 & 22.7 & 43.5 & 51.9 \\
  \hline
  ResNet50 (baseline)& \multirow{5}{4em}{RetinaNet} & 37.74M & 239.32 & 35.6 & 55.5 & 38.2 & 20.0 & 39.6 & 46.5 \\
  -Shared (ours)&  & \textbf{32.92}M & 239.12 & \textbf{36.2} & 55.0 & 38.6 & 20.3 & 39.7& 47.1 \\
  -Shared$^{\ddagger}$ (ours)&  & \textbf{30.67}M & 239.12 & 35.6 & 54.1 & 38.2 & 19.8 & 39.2 & 46.9 \\
  \cline{1-1}\cline{3-10}
  ResNet101 (baseline)&  & 56.74M & 315.39 & 37.7 & 57.5 & 40.4 & 21.1 & 42.2 & 49.5 \\
  -Shared (ours)& & \textbf{41.88}M & 315.09  & 37.7 & 56.8 & 40.3 & 21.2 & 41.5 & 49.5 \\
    \hline
  \end{tabular}
\end{footnotesize}
\end{table*}
\begin{table*}[thbp!]
\centering
\begin{footnotesize}
  \caption{Instance segmentation results using Mask R-CNN on COCO val2017.}
  \label{table:coco17segmentation}
  \begin{tabular}[tbhp]{l | l | l | l | l | l | l }
  \hline 
  Backbone & $AP$ & $AP_{50}$ & $AP_{75}$ & $AP_{S}$ & $AP_{M}$ & $AP_{L}$ \\
  \hline
ResNet50 (baseline) & 34.1 & 55.5 & 36.2 & 16.1 & 36.7 & 50.0 \\
-Shared (ours) & \textbf{34.5} & 55.4 & 36.9 & 18.9 & 37.7 & 46.5  \\
  \hline
  ResNet101 (baseline)& 35.9 & 57.7 & 38.4 & 16.8 & 39.1 & 53.6 \\
-Shared (ours)&  35.9 & 57.2 & 38.3& 19.1& 39.2& 48.6 \\
  \hline
  \end{tabular}
\end{footnotesize}
\end{table*}

Table \ref{table:coco17detection} shows the results on \texttt{val2017}
containing 5000 images.
Our backbone networks with shared filter bases 
consistently outperform baselines in all detectors 
in terms of COCO's standard metric AP 
while saving up to 28.1\% parameters.
In Table \ref{table:coco17segmentation}, 
our models also achieve similar performance improvement in instance segmentation 
using Mask R-CNN. 
This consistent performance improvement is obtained simply by replacing the backbone networks with ours
and it demonstrates that 
the proposed recursive convolution block design and the orthogonality regularization
enables learning better feature representations with a smaller amount of parameters.


\subsection{Analysis: Effect of Orthogonality Regularization}
\label{sec:analysis-ortho}
\begin{figure*}[tbhp!]
\centering
  \includegraphics[width=1.0\linewidth]{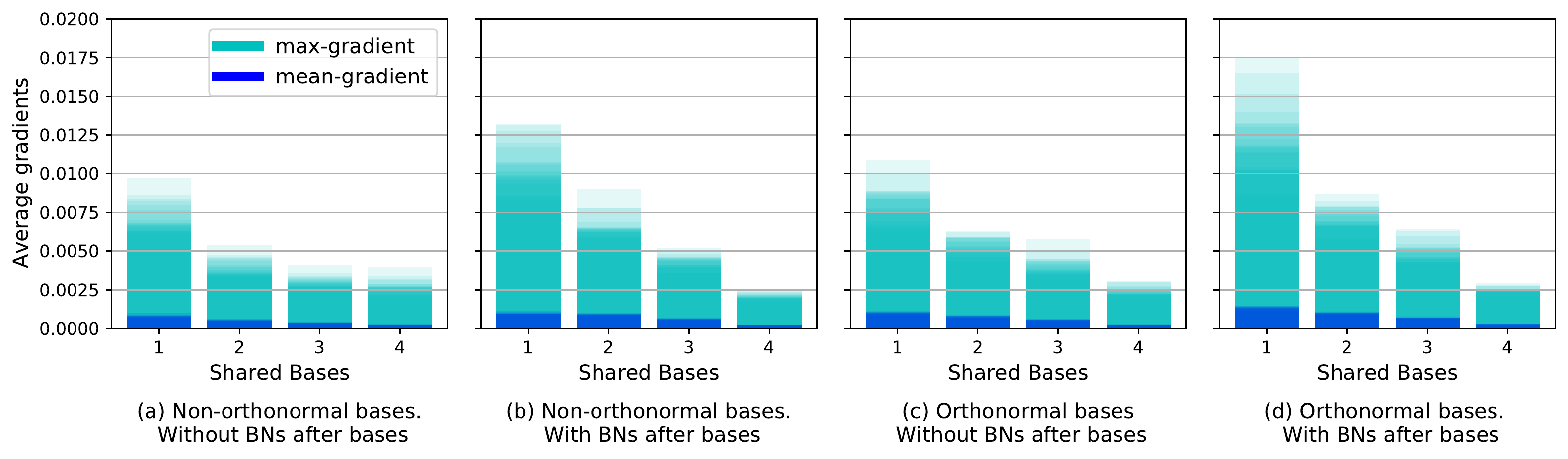}
  \caption{The flows of gradients in 4 shared filter bases of ResNet34-S16U1 at the same epoch.
  For comparison, the orthogonality regularization and the batch normalization (BN) following 
  the filter bases are turned on and off. 
  In (b) and (c), BNs and the orthogonality regularization, respectively, improve the flow of gradients.
  In (d), when both BNs and orthogonality regularization are applied together, 
  the strongest flow of gradients is observed. This trend is consistently observed 
  during the training.
  }
  \label{fig:gradientflow}
\end{figure*}

To investigate the effect of the orthogonality regularization during training,  
we track the flows of gradients while training ResNet34-S16U1.
Figure \ref{fig:gradientflow} shows the traces of the maximum and the mean gradients flowing in the filter bases 
during an epoch. 
On every 10 iterations of a batch, the maximum and the mean gradients are overlapped on top of the old plots. Therefore, the bars look darker if bars are more overlapped. 
\citet{jastrzebski2018residual} and \citet{guo2019dynamic}
showed that unshared batch normalization (BN) strategy mitigates
the vanishing/exploding gradients problem, 
and our result in Figure \ref{fig:gradientflow}-(b) shows 
that unshared BNs following shared filter bases 
improve the flow of gradients. 
When the proposed orthogonality regularization is applied  
to the shared filter bases, in Figure \ref{fig:gradientflow}-(c), 
similar effect is observed on gradients.
When both unshared BNs and the orthogonality regularization are applied together,
in Figure \ref{fig:gradientflow}-(d),
further stronger, but still bounded, flows of gradients are observed.
This trend is consistently observed during training.
We conjecture that this healthy flow of gradients 
enables learning better feature representations during optimization process, 
resulting in superior performance on various tasks.

\begin{figure*}[htbp!]
\centering
  \includegraphics[width=1.0\linewidth]{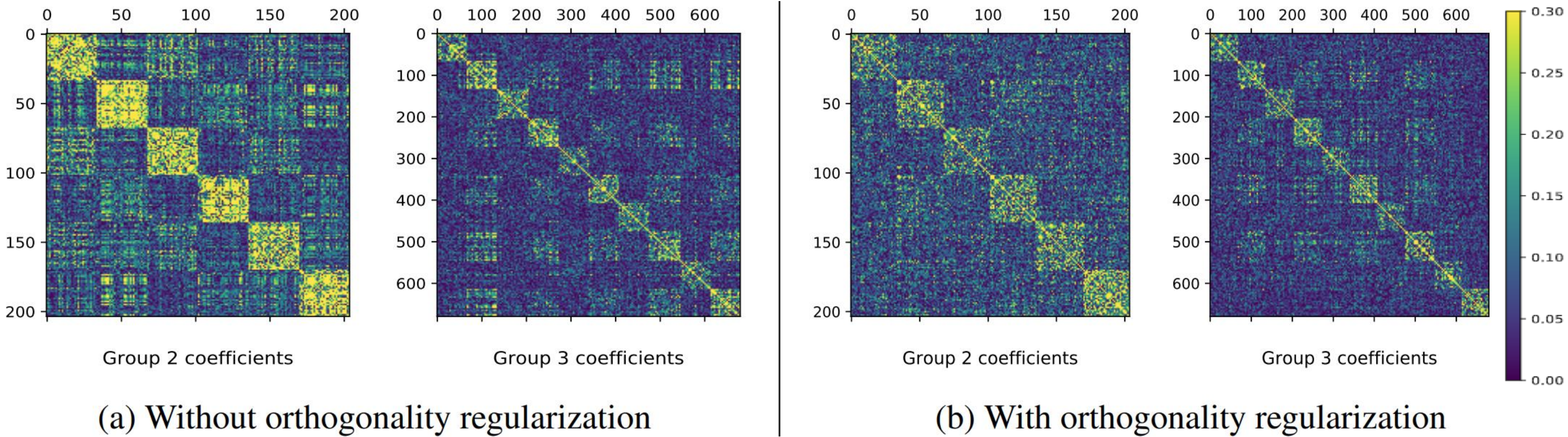}
  \caption{Cosine similarities of coefficients of the 2nd and the 3rd residual block groups in ResNet34-S16U1.
  X and Y axes are indexes to the coefficients of the residual block groups
  sharing filter bases. 
  Brighter colors indicate higher similarity.
  In (b), when the orthogonality regularization is applied, 
  the similarity is clearly lowered, implying less redundancy in parameters. 
  }
  \label{fig:cosine-similarity}
\end{figure*}

To further analyze the effect of the orthogonality regularization, in Figure \ref{fig:cosine-similarity},
we illustrate absolute cosine similarities of 
all coefficients of the 2nd and the 3rd residual block groups of ResNet34-S16U1.
The X and Y axes display the indexes to the coefficients
in the residual blocks. 
In Figure \ref{fig:cosine-similarity}-(b), 
we can clearly see that coefficients manifest 
lower similarities when the orthogonality regularization is applied. 
In Figure \ref{fig:cosine-similarity}-(a), when the orthogonality regularization
is not applied,
an interesting grid pattern is observed in Figure \ref{fig:cosine-similarity}-(a). 
This repetitive grid pattern might be related to ResNets' nature of 
iterative feature refinement \citep{jastrzebski2018residual}.
However, 
such high cosine similarity is directly related to the higher redundancy in the networks.
When our orthogonality regularization is applied, such repetitive patterns are less evident,
implying that the recursive convolution layers 
learn better feature representations
with less redundancy.

\section{Conclusions}
We introduce a recursive convolution block design
and effective training method for parameter-efficient CNNs.
In this work, a common filter basis, shared by repeating convolution layers,  
is learned while effectively avoiding
the vanishing/exploding gradients problem 
through the proposed orthogonality regularization. 
Experimental results on image classification and image detection 
show that our approach consistently outperforms
over-parameterized counterpart models while significantly saving parameters.
This consistent performance improvement
demonstrates that 
the proposed approach
enables effective learning of better feature representations
while a significant amount of parameters are shared.
We believe that the proposed recursive convolution blocks and training method
suggests important possibilities 
for neural architecture search (NAS) to explore resource-efficient CNNs.

\begin{ack}
We thank the anonymous reviewers for their constructive comments and suggestions.
This work was supported by the National Research Foundation of Korea (NRF) Grants 
Funded by the Ministry of Science and ICT under Grant NRF-2019R1F1A1060959.
\end{ack}

\bibliography{neurips_2021}
\bibliographystyle{neurips_2021}

\vfill

\pagebreak

\vfill

\pagebreak

\appendix
\section{Appendix}

\subsection{Effects of Ranks of Shared/Unshared Bases}
\label{sec:analysis-rank}
\begin{figure*}[htbp!]
\begin{footnotesize}
    \centering
    \includegraphics[width=1.0\linewidth]{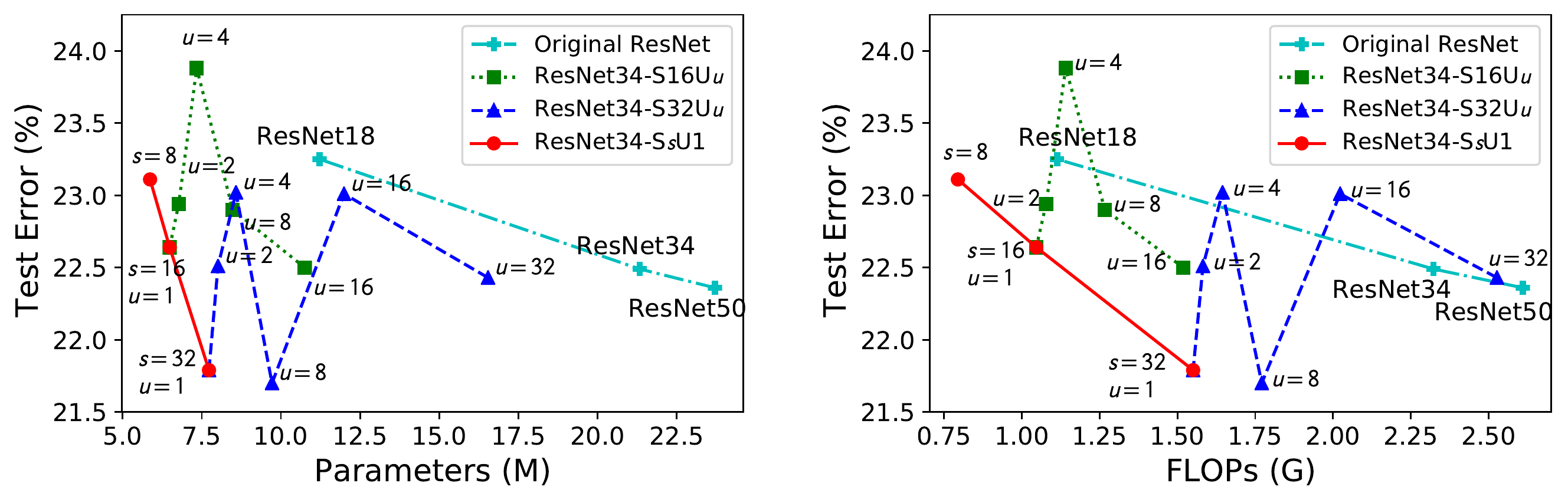} 
  \caption{Testing errors vs. the number of parameters and FLOPs on CIFAR-100.
  The number of shared basis elements (\textit{s}), and non-shared basis elements (\textit{u}) are varied.
  Using more shared basis elements results in better performance. 
  In contrast, using more non-shared elements does not always improve performance, implying the difficulty of training networks with larger parameters. 
  } 
  \label{fig:acc_params_flops}
\end{footnotesize}
\end{figure*}

Figure \ref{fig:acc_params_flops} shows test errors
on CIFAR-100 as parameters and FLOPs
are increased by varying the number of shared/non-shared filter basis elements
of networks. 
In general, the higher performance is expected with the more parameters. 
We observe that this presumption is true for shared basis elements.
For instance, when the number of shared basis elements $s$ is varied from 8 to 32, 
the test error sharply decreases from 23.1\% to 21.7\%.
However, non-shared basis elements manifest counter-intuitive results. 
Although a small number of non-shared basis elements (e.g., $u=1$) 
are clearly beneficial to the performance, 
the higher $u$'s do not always lead to the higher performance. 
For instance, when $u=4$, both ResNet34-S16U$u$ and ResNet34-S32U$u$
show the worst performance. 
This result demonstrates the difficulty of training networks with larger parameters. 
Further study is required for this problem.

\vfill


\end{document}